\documentclass[runningheads]{llncs}
\usepackage[T1]{fontenc}
\usepackage{graphicx,verbatim}
\usepackage{amsmath,amssymb,amsfonts}

\usepackage{algorithm}
\usepackage[noend]{algpseudocode}

\usepackage{hyperref}
\usepackage{marvosym}  

\usepackage{subcaption}
\usepackage{cite}
\usepackage{cleveref}
\usepackage{cleveref}
\crefname{table}{Table}{Tables}
\Crefname{table}{Table}{Tables}
\crefname{figure}{Fig.}{Figs.}
\Crefname{figure}{Fig.}{Figs.}

\begin{document}
\title{Graph-based LLM over Semi-Structured Population Data for Dynamic Policy Response}
%
\author{
Daqian Shi\inst{1}\textsuperscript{\Letter}
\and Xiaolei Diao\inst{2}
\and Jinge Wu\inst{1}
\and Honghan Wu\inst{1,3}
\and Xiongfeng Tang\inst{4}
\and Felix Naughton\inst{5}
\and Paulina Bondaronek\inst{1}\textsuperscript{\Letter}
}
\authorrunning{Daqian Shi et al.}

\institute{Institute of Health Informatics, UCL, UK 
\and
School of Electronic Engineering and Computer Science, QMUL, UK
\and
School of Health and Wellbeing, University of Glasgow, UK\\
\and
The Second Norman Bethune Hospital of Jilin University, China\\
\and
School of Health Sciences, University of East Anglia, UK\\
}

\maketitle             

\begin{abstract}
Timely and accurate analysis of population-level data is crucial for effective decision-making during public health emergencies such as the COVID-19 pandemic. However, the massive input of semi-structured data, including structured demographic information and unstructured human feedback, poses significant challenges to conventional analysis methods. Manual expert-driven assessments, though accurate, are inefficient, while standard NLP pipelines often require large task-specific labeled datasets and struggle with generalization across diverse domains. To address these challenges, we propose a novel graph-based reasoning framework that integrates large language models with structured demographic attributes and unstructured public feedback in a weakly supervised pipeline. The proposed approach dynamically models evolving citizen needs into a need-aware graph, enabling population-specific analyses based on key features such as age, gender, and the Index of Multiple Deprivation. It generates interpretable insights to inform responsive health policy decision-making. We test our method using a real-world dataset, and preliminary experimental results demonstrate its feasibility. This approach offers a scalable solution for intelligent population health monitoring in resource-constrained clinical and governmental settings.


\end{abstract}

\section{Introduction}
Timely and accurate analysis of population-level data is essential for effective decision-making during public health emergencies such as the COVID-19 pandemic \cite{chen2020artificial}. As governments and healthcare systems seek to understand the evolving needs and behaviors of citizens, they have to find evidence from vast volumes of semi-structured data stream \cite{shi2024leveraging}. This includes structured demographic records, such as age, gender, and socioeconomic indicators, and unstructured human feedback drawn from sources like community surveys, public helplines, and social media platforms \cite{shakeri2021digital}. These multimodal, asynchronous data streams hold potential insights for targeting interventions, allocating resources, and shaping policies \cite{consoli2025epidemiological}. However, traditional approaches for interpreting such complex data landscapes are increasingly proving inadequate in dynamic and resource-constrained environments.

Manual, expert-driven assessment pipelines are often considered the gold standard due to their domain-informed accuracy and context sensitivity. Yet, these processes are inherently slow, labor-intensive, and impractical for handling real-time or large-scale feedback during health crises \cite{sarker2024natural}. In contrast, automated natural language processing (NLP) systems can process unstructured data at scale but face significant limitations: they typically require task-specific labeled datasets, are rigid in adapting to new or domain-shifted content, and often fail to capture the nuanced and evolving concerns of heterogeneous populations \cite{alon2025leveraging}. Consequently, many valuable signals in public feedback go underutilized, leading to critical blind spots in healthcare response strategies, particularly when timely local insights are needed to protect vulnerable groups \cite{yang2021sudden}.

To address these challenges, we propose a graph-based large language model (LLMs) framework designed to integrate and analyze semi-structured population data in a dynamic, weakly supervised manner. At the core of our system is a hybrid framework that connects structured demographic attributes with unstructured human feedback through graph representations. By leveraging the reasoning capabilities of LLMs in conjunction with a need-aware graph, our method enables fine-grained, interpretable, and temporally adaptive analysis of public sentiment and needs. Noting that the construction of the graph adopts a pipeline that combines automatic knowledge extraction with expert validation. Therefore, the proposed method requires only minimal task-specific supervision, avoiding the high cost of large-scale manual annotation. Our contributions are threefold:
\begin{enumerate}
    \item We introduce a novel graph-based LLM analyzing framework that integrates structured and unstructured population data to support dynamic policy response.
    \item We develop a pipeline that allows for interpretable, need-aware insights with minimal reliance on manually labeled datasets.
    \item We run the proposed method on a real-world dataset as a case study, demonstrating its potential for effective population health monitoring in resource-constrained clinical and governmental contexts.
\end{enumerate}

\section{Related Work}

\subsection{Unstructured Population Data Analysis}

Early public-health surveillance relied on rule-based or lexicon-based sentiment engines such as VADER, which score text with handcrafted lexical heuristics and polarity intensifiers \cite{Hutto2014VADER}.  These systems are quick to deploy but brittle when confronted with concept drift. Traditional machine-learning pipelines that paired TF-IDF features with linear classifiers showed similar limitations on COVID-19 tweets \cite{BoonItt2020Twitter}.  Moreover, topic modeling techniques such as Latent Dirichlet Allocation (LDA) have been employed to identify major topics in large-scale pandemic-related discussions by clustering textual content to uncover the public’s primary concerns \cite{xue2020public}. Bondaronek et al. propose a human‑in‑the‑loop Structural Topic Modeling approach that can efficiently analyze free‑text feedback from 37,914 UK adults to extract actionable themes, offering rapid, scalable insights to inform improvements in the NHS Test \& Trace service \cite{bondaronek2023user}. However, systems employing unsupervised clustering algorithms still face several limitations, such as the continued need for manual effort to interpret and label the resulting clusters. Moreover, the labeling process tends to be subjective, making it challenging to generate in-depth and consistent analytical insights.

Simultaneously, lightweight n-gram logistic regression classifiers have proven effective in rapidly detecting symptom mentions or misinformation in real-time Twitter streams, offering a fast and reasonably accurate alternative under constrained computational resources \cite{didi2022covid}. Transformer models pretrained on outbreak corpora provide richer context.  COVID-Twitter-BERT (CT-BERT) is trained on 160 million pandemic-related tweets and improves classification accuracy by 10-30 percent over general BERT \cite{Mueller2020CTBERT}. Nevertheless, supervised fine-tuning remains label hungry, and domain shift degrades performance as vocabulary evolves. In summary, these limitations highlight a critical research gap: current approaches to analyzing public feedback often require predefined output targets and substantial manual annotation, yet there remains a lack of effective methods for generating deep insights without heavy reliance on human-labeled data.

\subsection{Graph-based LLM for Unstructured Data Analysis}

Recent studies have explored the integration of large language models (LLMs) with graph-based reasoning algorithms to enhance factual reliability and depth of inference across various free text datasets \cite{Yasunaga2021QAGNN, Wang2022KAdapter}. For instance, K-BERT incorporates knowledge graphs into BERT by injecting triples into the self-attention mechanism, using entity relationships to constrain contextual encoding and thereby strengthen reasoning capabilities \cite{Liu2020KBERT}. KEPLER jointly optimizes language modeling and knowledge embedding objectives, improving few-shot relation extraction without requiring additional fine-tuning data \cite{Wang2020KEPLER}. Extending this approach, CoLAKE constructs a lexical knowledge graph to unify entity-level and lexical-level contexts during masked language model training \cite{Sun2020CoLAKE}.

In the healthcare domain, DR.KNOWS retrieves UMLS knowledge paths as prompts for LLMs, significantly improving diagnostic F1 scores on electronic health record data \cite{Gao2023DRKnows}. MedRAG combines retrieval-augmented generation with a diagnostic knowledge graph to reduce misdiagnoses on the DDXPlus benchmark \cite{zhao2025medrag}. These systems demonstrate strong reasoning capabilities by leveraging LLMs’ understanding of unstructured data, further enhanced by the integration of structured knowledge graphs, which support deeper factual inference. Additionally, they offer potential to reduce annotation demands through semi-supervised approaches \cite{white2024informing}. However, adapting these techniques to the deep analysis of population-level needs requires further design considerations, such as integrating discussions with population-level metadata and addressing temporal dynamics to capture evolving public needs over time \cite{hall2022review}.

\section{The Proposed Method}

\begin{figure}[t]
  \centering
  \includegraphics[width=1\linewidth]{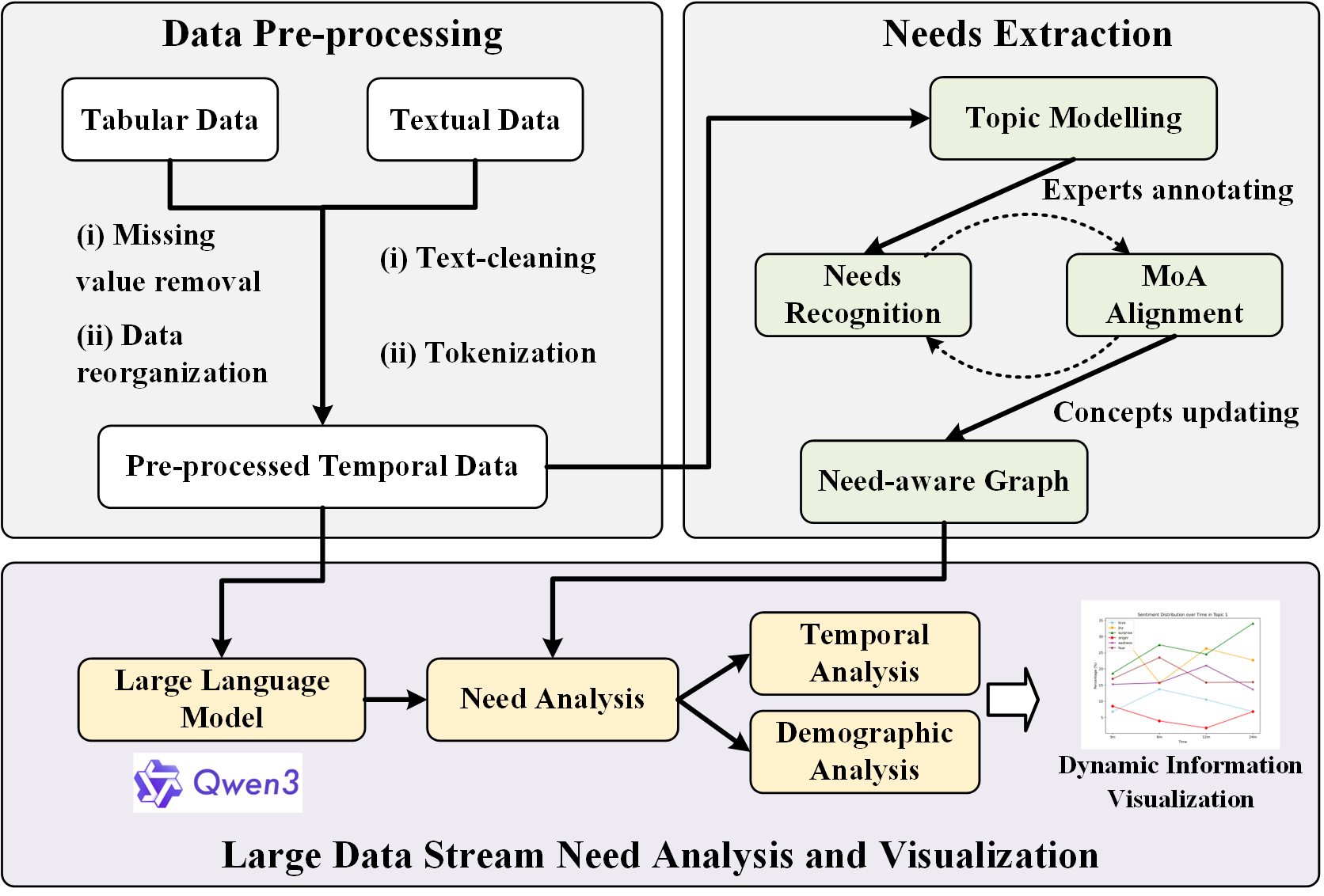}
  \caption{The overall demonstration of our proposed framework for analyzing semi-structured large population data-stream.}
  \label{fig:pipeline}
\end{figure}

This study introduces a novel graph-based LLM analyzing framework that integrates structured demographic features with unstructured public feedback in a weakly supervised pipeline. Citizen needs are dynamically modeled as a need-aware graph, enabling population-specific analysis along key features such as age, gender, and the Index of Multiple Deprivation (IMD). Figure \ref{fig:pipeline} depicts the complete framework, which is organized into three tightly coupled modules: (i) \textbf{data pre-processing}, (ii) \textbf{needs extraction}, and (iii) \textbf{large language-model need analysis and visualization}. Given streams of tabular (demographic) and textual (feedback) data, we first perform unified pre-processing to obtain a clean, time-stamped corpus. We then extract needs by identifying latent issue categories and aligning them with a library of Mechanism of Actions (MoA). The resulting signals populate an evolving need-aware graph. A lightweight local LLM consumes graph-enhanced inputs to score current needs. Finally, temporal and demographic analytics generate deep insights that are presented visually, thereby informing responsive health-policy decision-making. We detail each module accordingly in the rest of this section.

\subsection{Data Pre-processing}

Population-level studies of public health typically involve \emph{semi-structured} data: machine-readable demographic tables coexist with highly variable free-text feedback that captures citizens’ lived experience. This dual-stream design mirrors real-world surveillance settings in which numeric registries and narrative feedback should be analyzed jointly. We treat the demographic registry and the free-text corpus as two synchronized streams. For the \emph{structured} stream, missing values were either imputed or removed; records were anonymized and re-indexed by user ID. Age was separated into four bands (18-29, 30-39, 40-49, 50+), producing strata for subgroup analysis.  For the \emph{unstructured} stream, we executed a standard cleaning pipeline: removal of URLs, emojis, and boilerplate; language detection and lower-casing; sentence segmentation; Word piece tokenization; and stop-word elimination \cite{wu2023knowlab}. Each text entry was then linked back to its demographic vector and assigned to one of the four follow-up time points, producing a harmonized, time-stamped dataset that feeds the downstream graph construction module.

\subsection{Needs Extraction}
The pre-processed unstructured text is passed into the needs-extraction module, which identifies the demands articulated by citizens \cite{bondaronek2025public}, maps them to the behavioural-science ontology of MoA, and writes the results into a continuously updated need-aware graph that provides structural context for subsequent LLM reasoning. The workflow proceeds as follows.

\begin{itemize}
\item \textbf{Stage1: Topic modelling}. We apply Latent Dirichlet Allocation (LDA) \cite{blei2003latent} to the unlabelled corpus.  The number of topics is tuned to the setting that minimises held-out perplexity (typically around ten in our data).  LDA yields coarse themes, such as \emph{medicine shortage} or \emph{feelings of isolation}, without requiring manual annotation.

\item \textbf{Stage2: Needs recognition}. Domain experts inspect each topic's top tokens and representative documents, then assign a concise \emph{need label} drawn from a seed lexicon (e.g.\ \emph{hygiene needs}, \emph{food needs}, \emph{mental-health support}).  A lightweight rule set maps topic distributions to these labels, producing pseudo-tags that serve as weak supervision for recognition.

\item \textbf{Stage3: MoA alignment}. The labelled snippets are matched to our in-house MoA ontology\footnote{The Mechanisms of Action (MoA) ontology is a behaviour-science taxonomy that organizes intervention techniques by the psychological or environmental mechanisms through which they operate.}.  A local LLM Qwen-1.7B then proposes the most relevant MoA concept for each recognized need\footnote{The alignment can also be performed manually when expert oversight is required.}.  If a need has been encountered before, its dictionary entry is retained; otherwise, the newly detected need and its MoA mapping are appended to the seed lexicon. By updating the lexicon and graph in this online manner, the system continuously improves recognition quality while keeping the need-aware graph current as new data flows in.
\end{itemize}

We initiate a five-layer sub-graph and continuously enrich it as fresh information arrives.  Formally, at time step \(t\) we maintain:
\[
\mathcal{G}_{t} = \bigl(\, \mathcal{V}_{t},\; \mathcal{E}_{t} \bigr), 
\quad
\mathcal{V}_{t} = 
\mathcal{C} \;\cup\; \mathcal{N}_{t} \;\cup\; \mathcal{O}_{t} \;\cup\; \mathcal{B} \;\cup\; \mathcal{I},
\]
where the vertex sets correspond to the following five semantic layers:
\begin{itemize}
  \item \textbf{Category} \(\mathcal{C}\): high-level domains such as \emph{Need} or \emph{Obstacle}.  
  \item \textbf{Need} \(\mathcal{N}_{t}\): concrete demands extracted from text (e.g.\ \emph{food needs}).  
  \item \textbf{Obstacle} \(\mathcal{O}_{t}\): stated barriers that hinder fulfilment (e.g.\ \emph{no nearby site}).  
  \item \textbf{COM-B} \(\mathcal{B}\): behavioural determinants that link each obstacle to the Capability-Opportunity-Motivation model.  
  \item \textbf{BCIO class} \(\mathcal{I}\): Behaviour Change Intervention Ontology nodes specifying evidence-based techniques applicable to the need–obstacle pair.
\end{itemize}
Edges in \(\mathcal{E}_{t}\) encode hierarchical \textit{is-a} or \textit{belongs-to} relations and inherit
timestamps from the originating documents.  As new feedback streams in, stages 2 and 3 run online; the resulting node increments
\(\Delta\mathcal{V}_{t}\) and edge increments \(\Delta\mathcal{E}_{t}\) update the graph according to
\[
\mathcal{V}_{t+1} = \mathcal{V}_{t} \cup \Delta\mathcal{V}_{t},
\quad
\mathcal{E}_{t+1} = \mathcal{E}_{t} \cup \Delta\mathcal{E}_{t}.
\]
This incremental scheme keeps the graph current and supplies structural constraints for the downstream LLM, which draws on the graph to produce interpretable, population-related inferences.

\subsection{Dynamic Need Analysis and Visualization}
The outputs of the need-aware graph and LLM reasoning feed a \emph{dynamic analysis module} that quantifies how needs evolve over time and how they differ across population subgroups. Two complementary components are implemented. We deploy the locally hosted LLM Qwen 3 \cite{yang2025qwen3} as the core inference engine for this module. Its outputs drive the dynamic analysis, quantifying how needs shift over time and how they differ across population segments. The reasoning is constrained by the need-aware graph \cite{wu2024slava}, so the model concentrates on three aspects: identifying needs,  diagnosing their underlying causes, and suggesting potential solutions. The findings are delivered in two complementary formats to support dynamic policy response: (i) a concise natural-language report that summarizes the key needs and recommended interventions, and (ii) visual dashboards that display the quantifiable elements, such as need prevalence and sentiment trajectories, in a graphical form.

\noindent\textbf{Temporal analysis}.
For every time window \(t\) we compute the prevalence of a need \(n\) as:$P_{n,t}= \frac{\operatorname{count}(n,t)}{\sum_{n'}\operatorname{count}(n',t)}$, where \(\operatorname{count}(n,t)\) is the number of documents in which the LLM assigns need \(n\) the highest score.  Plotting \(P_{n,t}\) as a line chart reveals shifts in public attention. For instance, early in the pandemic, peaks appear for \textit{virus transmission} and \textit{personal protection}, whereas later windows highlight \textit{employment insecurity} and \textit{family dynamics}. We also attach sentiment trajectories by running a BERT-based emotion classifier fine-tuned on social-media text \cite{hoang2019aspect}, allowing us to observe whether concern about, e.g., remote schooling is predominantly negative or neutral as the crisis unfolds.

\noindent\textbf{Demographic analysis}.
The same need scores are stratified by demographic attributes (age band, gender, IMD decile).  For each subgroup \(d\) we compute \(P_{n,t}^{(d)}\) analogously and visualise disparities.  For instance, we find that women report \textit{mental-health support} needs 1.4 times more frequently than men during the six-month follow-up, while low-IMD respondents exhibit persistently lower \textit{living satisfaction} concerns.

\section{Experimental Observations}

We test the proposed framework via a workable case study, drawing on a longitudinal corpus collected from 1,045 UK residents during the COVID-19 pandemic \cite{bondaronek2023user}.  The dataset spans twenty-four months and couples structured descriptors, age, gender, and the Index of Multiple Deprivation (IMD), with repeated, open-ended survey responses.  Each participant contributed text at four follow-up points (3, 6, 12, 24 months), yielding 3,812 timestamped documents.  Our main observations are summarized below.

\subsection{Temporal Analysis}

Applying LDA to the 3,812 time-stamped responses revealed five coherent thematic clusters that recur throughout the corpus: (1) \textit{Mental Health and Emotions}, (2) \textit{Physical Health and Behaviours}, (3) \textit{Economy and Work}, (4) \textit{Coping Strategies and Positive Behaviours}, and (5) \textit{Constraints and Control}.  Although these themes persisted across the full 24-month observation window, their relative prevalence changed substantially from wave to wave:

\begin{itemize}
\item \textbf{0-3 months.} Many respondents expressed concerns about infection and difficulties obtaining essential supplies. At the same time, individuals began to mention topics such as online meetings and the purchase of virtual exercise classes. These accounts reflect residents’ psychological anxiety and sense of lack of control during the early phase of the pandemic.

\item \textbf{6 months.} As the healthcare system stabilized, discussions shifted towards employment instability and income loss, highlighting the escalating economic impact of the pandemic. Mentions of business closures and furlough anxieties increase. On a positive note, respondents also talked about spending time with family and engaging in outdoor physical activities.

\item \textbf{12-24 months.} In the medium and longer term, discussions about isolation and restrictions gradually decreased. Some narratives focused on family gatherings, fitness routines, and healthy eating. Explicit fears of infection and anxiety about shortages of medical resources diminished over time; however, concerns related to mental health remained a prominent topic.
\end{itemize}

\noindent Sentiment trajectories exhibited a comparable but more nuanced arc. The BERT-based classifier indicated a sharp spike in negative affect, fear, anger, and confusion during the first survey wave; this declined steadily as respondents acclimatized to public-health measures, only to rebound when extended restrictions and economic setbacks compounded stress six months later. Conversely, positive sentiment showed a modest uptick during the strictest lockdown, fueled by reports of mutual aid, online social events, and the adoption of new coping routines such as home-exercise challenges. By the final wave, the corpus was dominated by neutral or emotionally fatigued language, phrases expressing resignation, boredom, or “pandemic burnout”, signalling widespread habituation to the emerging “new normal” and a dampening of both positive and negative emotional extremes.

\subsection{Demographic Analysis}

\begin{itemize}
\item \textbf{Gender.} Women reported \textit{Health and Emotional Stress} needs more frequently than men, especially in the first 6 months, the topic related to caregiving duties is also mentioned more frequently; men more often highlighted work-related stress and financial uncertainty.
\item \textbf{Age.} Respondents aged 18–29 exhibited the highest anxiety over disrupted education and career prospects, with a pronounced spike in social-relationship concerns at months 3 and 12. The 50\,+ cohort mentions more worries about accessing routine healthcare for chronic conditions at months 6 and 24.  
\item \textbf{Socioeconomic status (IMD).} Participants in the 1-5 IMD consistently expressed elevated stress over employment and basic needs; their references to \textit{Economy and Work} are higher than those with the higher-IMD groups.
\end{itemize}

\noindent These subgroup patterns confirm that COVID-19 has magnified pre-existing social inequalities. Mental-health trajectories and perceived needs are conditioned not only by the crisis timeline but also by intersecting demographic factors, underscoring the value of our need-aware, graph-constrained analysis for targeted policy response.

\section{Conclusion and Future Work}
Our study demonstrates that a graph-enhanced, weakly supervised LLM pipeline can capture the temporal evolution and demographic heterogeneity of pandemic-related needs and emotions, offering actionable insights for responsive public-health planning.  By linking unstructured feedback to a behaviour-science ontology and visualising subgroup trajectories, we show that citizen concerns shift from acute infection anxiety to longer-term psychosocial and economic challenges, with marked variations by age, gender, and socioeconomic status.  These findings underscore the importance of continuously monitoring population-level feedback to support timely, equitable interventions and to design policies that are both adaptive and inclusive. However, it should be noted that we also identified traces of inherent stereotypes in the results, which reveal certain intrinsic biases in AI-driven population analysis. While this demonstrates the capabilities of the proposed approach, it also highlights the limitations of the current work. Since the primary focus of our study is to establish a robust pipeline for analyzing user needs and perspectives, further validation and refinement of the results through domain experts will become the main focus of our subsequent work.

Our future work will focus on several directions. On the one hand, we plan to leverage large, expert-validated knowledge graphs to further strengthen behavioral and psychological interpretations of user needs and actions. On the other hand, we will further explore human-in-the-loop paradigms, emphasizing how domain experts, following initial analyses, can generate deeper and more robust insights, thereby enhancing the interpretability of analytical outcomes.

\begin{credits}
\subsubsection{\ackname} 
This research is supported by the Wellcome Trust (Project No.300252/Z/23/Z). This work was also supported by the UK’s Medical Research Council (Project No. MR/S004149/1, MR/X030075/1).

\subsubsection{\discintname}
The authors have no competing interests to declare that are relevant to the content of this article.

\end{credits}

\bibliographystyle{splncs04}
\bibliography{references}

\end{document}